\documentclass{ecai} 

\usepackage{graphicx}
\usepackage{booktabs}
\usepackage{wrapfig}

\usepackage{graphicx}
\usepackage{amsmath}
\usepackage{amsthm}
\usepackage{booktabs}
\usepackage{algorithm}
\usepackage{algorithmic}
\usepackage[switch]{lineno}

\usepackage{tabularx}
\usepackage[table]{xcolor}
\usepackage{array}
\usepackage{amssymb}
\usepackage{listings}
\usepackage{xcolor}

\newcommand{\CIFARSmall}{95.16}
\newcommand{\CIFARLarge}{76.48}
\newcommand{\ImageNetSmall}{ 43.42 }
\newcommand{\toolName}{BioNAS}

\newcommand{\BibTeX}{B\kern-.05em{\sc i\kern-.025em b}\kern-.08em\TeX}

\begin{document}


\begin{frontmatter}


\paperid{123} 


\title{Neural Architecture Search with Mixed \\ Bio-inspired Learning Rules}


\author[AB]{\fnms{Imane}~\snm{Hamzaoui}
\thanks{Corresponding Author. Email: ji\_hamzaoui@esi.dz.}}
\author[B]{\fnms{Riyadh}~\snm{Baghdadi}
}

\address[A]{École nationale Supérieure d'Informatique Algiers}
\address[B]{New York University Abu Dhabi}


\begin{abstract}
Bio-inspired neural networks are attractive for their adversarial robustness, energy frugality, and closer alignment with cortical physiology, yet they often lag behind back-propagation (BP) based models in accuracy and ability to scale. We show that allowing the use of different bio-inspired learning rules in different layers, discovered automatically by a tailored neural-architecture-search (NAS) procedure, bridges this gap. Starting from standard NAS baselines, we enlarge the search space to include bio-inspired learning rules and use NAS to find the best architecture and learning rule to use in each layer. We show that neural networks that use different bio-inspired learning rules for different layers have better accuracy than those that use a single rule across all the layers.
The resulting NN that uses a mix of bio-inspired learning rules sets new records for bio-inspired models: 95.16\% on CIFAR-10, 76.48\% on CIFAR-100, 43.42\% on ImageNet16-120, and 60.51\% top-1 on ImageNet. In some regimes, they even surpass comparable BP-based networks while retaining their robustness advantages. Our results suggest that layer-wise diversity in learning rules allows better scalability and accuracy, and motivates further research on mixing multiple bio-inspired learning rules in the same network.
\end{abstract}

\end{frontmatter}


\section{Introduction}
Bio-inspired neural networks have gained significant traction recently. They are more biologically plausible, more robust to adversarial attacks~\cite{sanfiz2021benchmarkingakrout,moraitis2022softhebb}, and have higher energy efficiency~\cite{malcolm2023comprehensivereviewspikingneural}. Despite their significant success, these methods are still in development and fall short in terms of accuracy and in scaling effectively for complex tasks. In this paper, we aim to use neural architecture search to improve state-of-the-art bio-inspired neural networks.

Neural Architecture Search (NAS) is a method that automatically searches for neural network architectures and effectively finds high-quality architectures. This is usually done by exploring different types of architectures and picking one that best satisfies an objective (e.g., minimizing the loss). Early work based on Evolutionary Algorithms (EAs) and Reinforcement Learning (RL) was able to find interesting architectures, however such methods take significant amount of time. AmoebaNet-B~\cite{real2019regularized-amoeba}, for example, takes around 3150 GPU days. Several efficient NAS methods have been developed since then, offering a trade-off between search time and accuracy. 
Differentiable Architecture Search (DARTS)~\cite{liu2018darts} is a notable example. It has shown success in designing differentiable neural architecture search.
EG-NAS~\cite{EG-NAS24}, a follow-up work, improved on top of DARTS using genetic algorithms. It has shown an ability to find an accuracy competitive with DARTS in 0.1 GPU days compared to 0.4 GPU day for DARTS on CIFAR10. 

In this paper, we build on top of DARTS and EG-NAS to propose \toolName{}, a framework for neural architecture search that explores different bio-inspired neural network architectures and learning rules. 
To the best of our knowledge, \toolName{} is the first NAS framework that explores the use of different bio-inspired learning rules in the same neural network. This choice is motivated by recent research that shows that the brain might be using different learning rules instead of a single rule~\cite{marblestone-2016}. 
We show that interesting architectures can be discovered within this framework, surpassing state-of-the-art bio-inspired neural networks,
and achieving an accuracy of \CIFARSmall{}\% on CIFAR10 (compared to 91.8\% for ResNet56 trained with Uniform Sign-concordant Feedbacks (uSF)), \CIFARLarge{}\% on CIFAR-100 (compared to 73.04\% on SNASNet), \ImageNetSmall{}\% on ImageNet16-120 and 60,51\% on ImageNet. In addition to the improved accuracy, the use of bio-inspired learning rules also leads to better robustness against adversarial attacks. We show that a part of this improvement comes from the use of different learning rules instead of using a single learning rule for all the layers.

The contributions of this paper are as follows
\begin{enumerate}
    \item We propose \toolName{}, a framework for neural architecture search designed for bio-inspired neural networks. To the best of our knowledge, \toolName{} is the first framework that incorporates learning rules as part of the neural architecture search.
    \item We show that using different bio-inspired learning rules improves the accuracy of bio-inspired models as well as their adversarial robustness.
   \item We release \toolName{} to the community. and make the code available   via this link (\url{https://anonymous.4open.science/r/LR-NAS-DFE1})
    .
\end{enumerate}

\vspace{-0.5cm}
\section{Background \& Related Works}
\subsection{Biologically Inspired Neural Networks}


Biologically inspired neural networks are neural networks that draw inspiration from principles observed in natural nervous systems. Broadly, these networks integrate biological insights in two ways:

\textbf{(i) Architectures.} This involves designing network structures that closely resemble biological neural circuits. Examples include using spiking neurons (which mimic biological neurons that communicate via discrete spikes), dendritic branching (reflecting the complex input-processing capabilities of real neurons), local recurrent connections (small-scale loops found abundantly in brain circuits), or sparsely connected layers similar to the olfactory circuits observed in fruit flies. Such architectures can still leverage conventional neural network optimization methods for training (e.g., back-propagation).

\textbf{(ii) Learning rules.} Alternatively, networks may retain standard deep learning architectures, such as convolutional neural networks (CNNs) or transformers, but adopt biologically plausible learning rules. These rules typically rely only on local information flow (each neuron updates based only on its immediate neighbors or local environment) or utilize random feedback connections. Such approaches avoid the strict symmetry and global knowledge required by traditional back-propagation algorithms.

Our framework specifically explores this second approach, extending Neural Architecture Search (NAS) to automatically select biologically plausible learning rules for each layer during network construction.
\vspace{-1em}

\subsection{Bio-inspired Learning Rules}
Back-propagation (BP)~\cite{rumelhart-1986} is the most widely used algorithm for training neural networks.  
It operates by first performing a forward pass, where the network processes inputs to produce predictions, and then executing a backward pass, computing gradients of the loss function to update network weights. However, BP relies on symmetric weight matrices—meaning the weights used for feedback (gradient calculation) are the transposed version of those used for the forward pass—a requirement considered biologically implausible.


To address this, several alternatives have been proposed.  
\emph{Feedback Alignment (FA)}~\cite{lillicrap-2016-feedback-alignement} simplifies the feedback process by sending random error signals from the network's output layer back to the hidden layers. In other words, it replaces the transposed weight matrix with a fixed random feedback matrix, relaxing the need for symmetry.
\emph{Direct Feedback Alignment (DFA)}~\cite{nokland2016direct} goes further by sending random feedback from the output layer directly to each hidden layer, bypassing intermediates.  
Further research indicates that using \emph{sign-concordant} feedback matrices—such as Uniform Sign Feedback (uSF) and Batchwise Random-Magnitude Sign Feedback (brSF)~\cite{liao2016important_usf_brsf}—can more closely mimic BP performance by preserving the signs of feedback signals.
Figure~\ref{fig:different-credit-assignment} summarizes these credit-assignment schemes. Other forward-only paradigms include \emph{PEPITA}~\cite{dellaferrera2023errordriven}, which uses a second forward pass modulated by error signals, and \emph{Forward-Forward (FF)}~\cite{hinton2022forwardforward}, which relies on a layer-wise “goodness” metric computed for positive and negative data. Finally, \emph{Hebbian learning}, inspired by synaptic plasticity (“cells that fire together wire together”), provides a local and biologically plausible update based on neuron co-activation. While early implementations were too simplistic for large-scale tasks, recent works~\cite{lagani_hebbian_conv,lagani_hebbian_imagenet} scale Hebbian rules to CIFAR-10 and ImageNet; Winner-Take-All mechanisms~\cite{journehebbian} further enhance scalability by dynamically adapting learning rates.

While these biologically plausible learning rules do not match the performance of backpropagation in terms of accuracy, recent work~\cite{sanfiz2021benchmarkingakrout} benchmarked different architectures trained with different learning rules on several datasets. The paper also highlighted their intrinsic robustness to adversarial attacks by a thorough evaluation on both white box and black box adversarial attacks.

\subsection{Neural Architecture Search (NAS)}
Neural Architecture Search (NAS) aims to automate the design of neural networks by optimizing both the architecture and its parameters. Differentiable Architecture Search (DARTS)~\cite{liu2018darts} introduces a gradient-based approach to NAS, enabling efficient exploration of large search spaces. By relaxing the discrete search space into a continuous one, DARTS allows for optimization via gradient descent, significantly reducing computational costs.

Building on DARTS, EG-NAS~\cite{EG-NAS24} incorporates evolutionary strategies like Covariance Matrix Adaptation Evolution Strategy (CMA-ES)~\cite{hansen2016cma} to enhance exploration and mitigate convergence to local optima. This hybrid approach alternates between gradient descent for network weights and evolutionary updates for architecture parameters, achieving better scalability for complex search spaces.

Recent efforts have explored robust NAS techniques, such as RobustNAS~\cite{zhu2023robustNAS}, which integrate adversarial robustness metrics into the NAS objective function. However, these methods often require expensive adversarial evaluations during the search. Our work differs by focusing on bio-inspired neural networks, leveraging their inherent robustness to noise and attacks without requiring explicit adversarial evaluations.

Other work~\cite{kim-2022-nas-snn,YAN2024ENASSNN} has previously used NAS for Spiking Neural Networks, a biologically inspired architecture used in computational neuroscience and event-based applications, they considered both forward and backward connections as part of their search space but didn't consider learning rules, which we do in this framework.

\subsection{Adversarial Attacks}
Adversarial attacks are malicious perturbations designed to fool neural networks, eroding their predictions. These attacks can be categorized as white-box (access to model parameters) or black-box (limited access). Common methods include Fast Gradient Sign Method (FGSM)~\cite{goodfellow2014explaining}, Projected Gradient Descent (PGD)~\cite{madry2017towards}, and Auto Projected Gradient Descent (APGD)~\cite{Croce2020}. FGSM computes perturbations in the direction of the gradient, while PGD iteratively applies smaller perturbations. APGD enhances efficiency by adapting step sizes during optimization.

Attacks like One-Pixel~\cite{su2019one} target only a few pixels in the input image to deceive the network while maintaining visual integrity. Recent studies~\cite{wu2024robustICLR} benchmark adversarial robustness in NAS-generated architectures, highlighting the susceptibility of BP-trained models to gradient-based attacks. Our exploration of bio-inspired NAS has the advantage of addressing this vulnerability by leveraging the intrinsic robustness of biologically plausible learning rules.

\section{Methods}

In this section, we introduce the NAS problem, outline the existing methods used in our framework, and describe the various phases and additions made during the development of our approach. We build upon DARTS and EG-NAS frameworks, incorporating both conventional backpropagation techniques and biologically plausible learning rules. Specifically, we focus on feedback alignment methods and other bio-inspired paradigms.

\subsection{Neural Network Framework}
We consider a neural network with an input vector \(\mathbf{z}\), output vector \(\mathbf{y}\), and activation function \(\phi(\cdot)\). The weight update in conventional backpropagation is derived as:

\begin{equation}
    \delta W_i = -\eta \frac{\partial L}{\partial W_i} = -\eta \left( \frac{\partial L}{\partial \mathbf{y}} \cdot \frac{\partial \mathbf{y}}{\partial \mathbf{z}_i} \cdot \frac{\partial \mathbf{z}_i}{\partial W_i} \right),
\end{equation}

where \(\eta\) is the learning rate, \(L\) is the loss function, and \(\mathbf{z}_i\) represents the pre-activation at layer \(i\). Bio-inspired methods modify this rule by replacing gradients with biologically plausible approximations, detailed as follows:
\begin{figure}
    \centering
    \includegraphics[width=\linewidth]{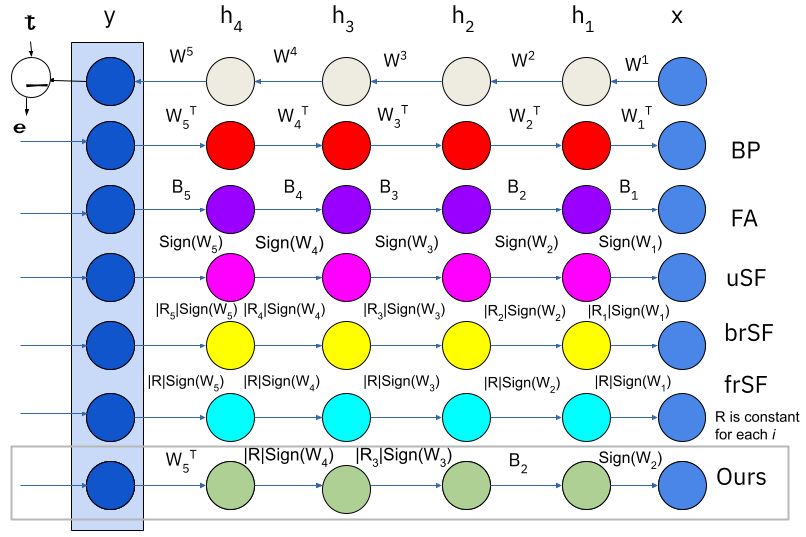}
    \vspace{1em}
    \includegraphics[width=0.6\linewidth]{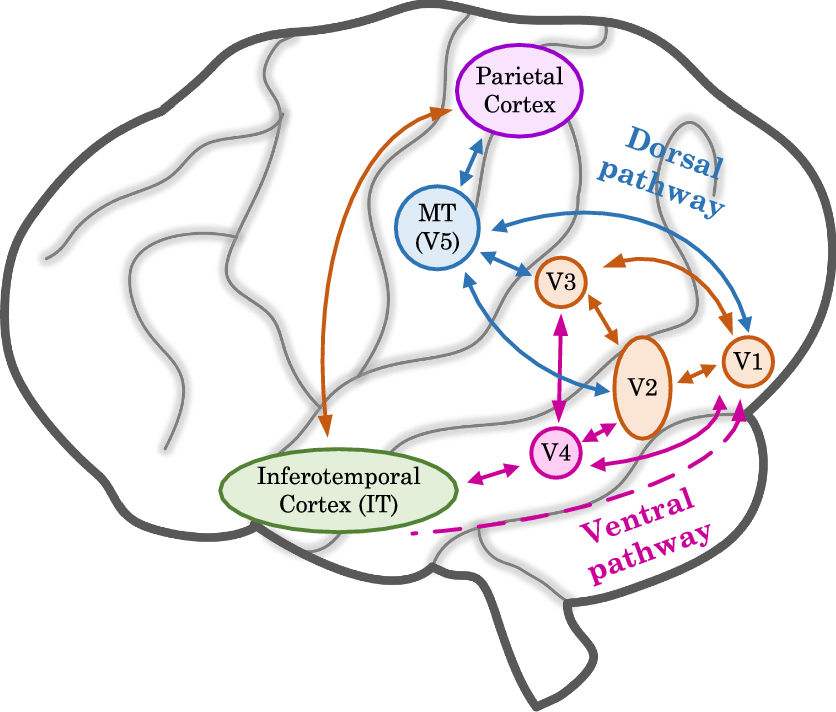}
    \vspace{-2em}
    \caption{
        \textbf{Top:} Conceptual depiction of different credit assignment techniques such as Backpropagation (BP) and Feedback Alignment (FA) across layers. 
        \textbf{Bottom:} Brain-inspired schematic showing inter-regional signaling pathways. A large number of feedforward and feedback connections exist in ventral and dorsal pathways of the visual cortex, even within the same cortex. Figure from \cite{Brain-inspired-figure}.
    }
    \label{fig:different-credit-assignment}
\end{figure}
\paragraph{Feedback Alignment (FA):} 
\begin{equation}
    \delta W_i = -\eta \left( \mathbf{e}_{i+1} \cdot \mathbf{B}_i \cdot \frac{\partial \mathbf{z}_i}{\partial W_i} \right),
    \label{formula1}
\end{equation}
where \(\mathbf{e}_{i+1}\) is the error at the next layer, and \(\mathbf{B}_i\) is a fixed random feedback matrix.

All of the remaining bio-inspired learning methods use the previous formula (formula \ref{formula1}) to update weights, except that they compute $\mathbf{B}_i$ in a different way.

\paragraph{Uniform Sign-concordant Feedbacks (uSF):} 

\begin{equation}
    \mathbf{B}_i = \text{sign}(\mathbf{W}_i),
\end{equation}
which transports the sign of the forward weights assuming unit magnitude.

\paragraph{Batchwise Random Magnitude Sign-concordant Feedbacks (brSF):} 
\begin{equation}
    \mathbf{B}_i = |\mathbf{R}_i| \cdot \text{sign}(\mathbf{W}_i),
\end{equation}
where \(|\mathbf{R}_i|\) is a random magnitude redrawn after each update.

\paragraph{Fixed Random Magnitude Sign-concordant Feedbacks (frSF):} 
\begin{equation}
    \mathbf{B}_i = |\mathbf{R}| \cdot \text{sign}(\mathbf{W}_i),
\end{equation}
where \(|\mathbf{R}|\) is a fixed magnitude initialized at the start of training.

\subsection{Search Space Design}
We extend the DARTS and EG-NAS frameworks by incorporating bio-inspired learning rules into the search space. The candidate operations include:

\begin{itemize}
  \item Separable convolutions (\(3 \times 3\) and \(5 \times 5\)) that can use FA, uSF, brSF, and frSF as learning rules.
  \item Dilated convolutions (\(3 \times 3\) and \(5 \times 5\)) that can use FA, uSF, brSF, and frSF as learning rules.
  \item \(3 \times 3\) max pooling and average pooling.
  \item Skip connections with feedback modes uSF, brSF, and frSF.
  \item Zero (no operation, used in DARTS).
\end{itemize}

In addition, we explored two biologically inspired convolutional mechanisms:

\begin{itemize}
    \item \textbf{Hebbian Convolution:} Inspired by Hebbian learning rules, we implement a convolutional layer that updates its weights using the outer product of normalized input and output patches, similar in spirit to~\cite{miconi2018differentiable}. During the forward pass, weights are iteratively adjusted using an outer product of normalized input and output patches, scaled by a small learning rate. This design aligns well with biological plausibility and prevents weight explosion through optional normalization. While it did not provide notable accuracy gains, it demonstrated the adaptability of our framework for diverse learning mechanisms.
\item \textbf{Predictive Coding Convolution:} Inspired by predictive coding, this convolution iteratively reduces prediction errors, with optional adaptive updates, gating, and normalization for stability.

\end{itemize}

\subsection{\toolName{}-DARTS and \toolName{}-EG}
\label{subsec:our_search}

Our NAS method explores a new search space that allows selecting not only architectural components (like convolutional blocks or pooling layers) but also different biologically-inspired learning rules for each part of the network. This section describes how we integrated this into two widely-used NAS frameworks: DARTS (gradient-based) and EG-NAS (evolutionary-based).

\paragraph{\toolName{}-DARTS (gradient–based).}
We follow the Differentiable Architecture Search (DARTS)~\cite{liu2018darts} formulation, where the goal is to find both the architecture and its parameters by solving a bi-level optimization problem.  In simple terms, we want to find an architecture (represented by $\boldsymbol{\alpha}$) that performs well on unseen data (validation set), while still being trained on the available training set. At a high level, the search alternates between two objectives:

\begin{equation}
    \min_{\boldsymbol{\alpha}}\; \mathcal{L}_{\mathrm{val}}\!\bigl(\mathbf{w}^{\star}(\boldsymbol{\alpha}),\boldsymbol{\alpha}\bigr)
\end{equation}
\noindent and
\begin{equation}
    \mathbf{w}^{\star}(\boldsymbol{\alpha})=\arg\min_{\mathbf{w}}\mathcal{L}_{\mathrm{train}}(\mathbf{w},\boldsymbol{\alpha}),
\end{equation}

\noindent where:
\begin{itemize}
    \item $\mathcal{L}_{\mathrm{train}}$ and $\mathcal{L}_{\mathrm{val}}$ are the training and validation losses.
    \item $\mathbf{w}$ represents the model weights and $\mathbf{w}^\star$ denotes the optimal weights that minimize the training loss.
    \item $\boldsymbol{\alpha}$ are continuous parameters (logits) that determine the architecture.
\end{itemize}


In practice, this optimization is approximated with a one-step update of weights and architecture parameters:
\begin{align}
\mathbf{w} &\leftarrow \mathbf{w} - \eta_{w}\nabla_{\mathbf{w}}\mathcal{L}_{\text{train}}, \\
\boldsymbol{\alpha} &\leftarrow \boldsymbol{\alpha} - \eta_{\alpha}\nabla_{\boldsymbol{\alpha}}\mathcal{L}_{\text{val}}.
\end{align}
where $\eta_{w}$ and $\eta_{\alpha}$ are the learning rate for $w$ and $\alpha$ respectively.

\begin{figure}[h]
    \centering
    \includegraphics[width=0.8\linewidth]{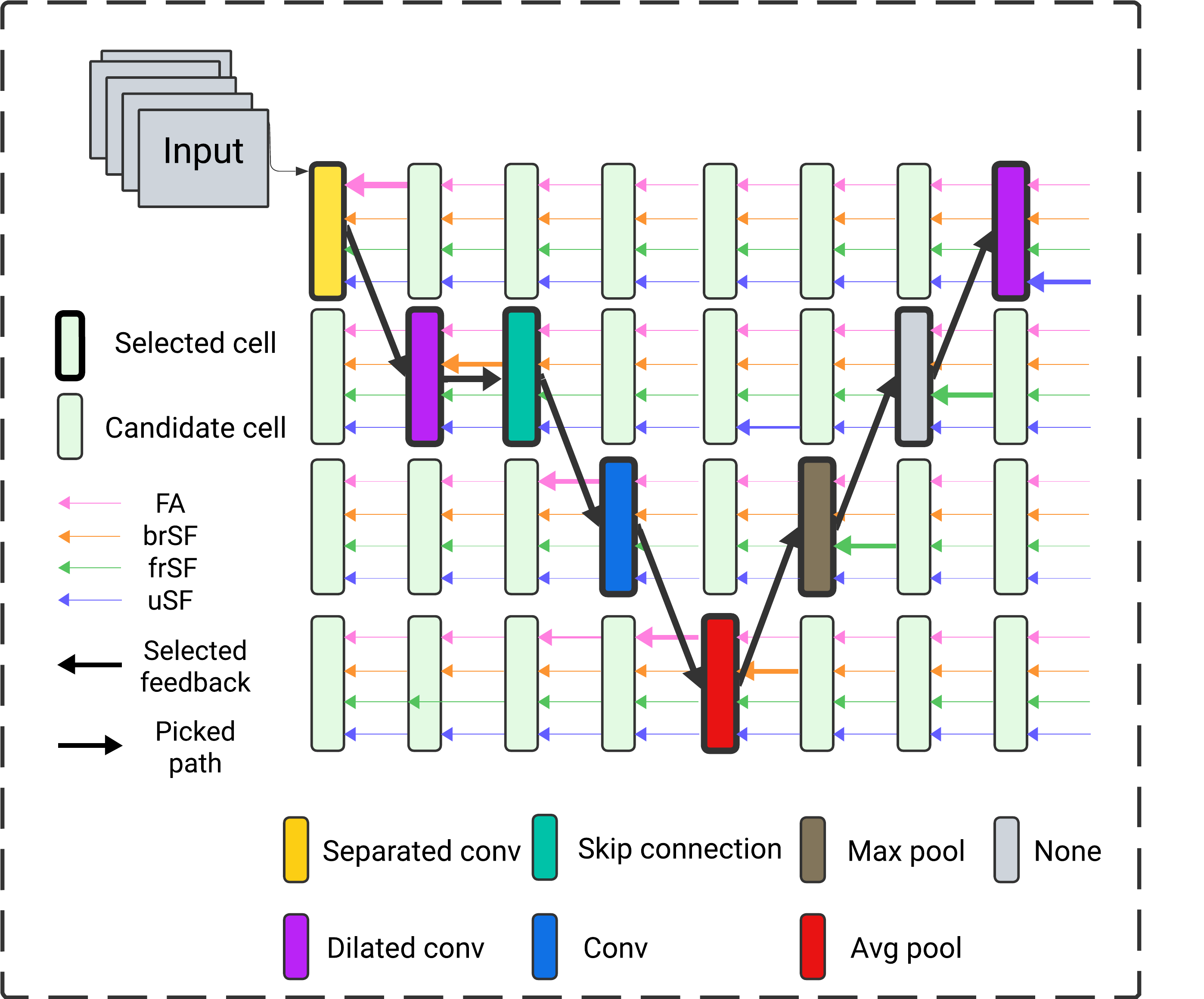}
    \vspace{-1em}

    \caption{Search process: each connection (edge) in the supernetwork selects the best candidate operation and associated learning rule. The backward arrow indicates the chosen learning rule (feedback mechanism).}
    \label{fig:search_process}
\end{figure}

\textbf{Supernetwork and search space.}  
DARTS defines a \emph{supernetwork} as a directed acyclic graph where nodes represent intermediate feature maps, and edges represent operations (e.g., convolutions, skip connections) applied between them. Each edge $e = (i \to j)$ is assigned a weighted combination of $K$ candidate operations.

In our framework, each candidate operation is a \textbf{pair} $(o_k, r_k)$ combining a computational block $o_k$ (e.g., $3{\times}3$ convolution) and a learning rule $r_k$ (e.g., FA, uSF, brSF, Hebbian). Let $K$ be the total number of operation-rule pairs per edge and $k \in \{1, \dots, K\}$ index them. The output of node $j$ is:

\begin{equation}
    \mathbf{x}_j \;=\; \sum_{i<j}\;\sum_{k=1}^{K}
          \frac{\exp(\boldsymbol{\alpha}_{e,k})}{\sum_{k'}\exp(\boldsymbol{\alpha}_{e,k'})}
          \;\; r_k\!\bigl(o_k(\mathbf{x}_i)\bigr).
    \label{eq:mixed_op}
\end{equation}

This formulation allows each edge to jointly select both a computation and its corresponding learning rule. After the search, the highest-probability $(o_k, r_k)$ pair on each edge is selected to form the final discrete architecture.

\vspace{1em}
\paragraph{\toolName{}-EG (evolutionary).}
To show the generality of our search space, we also integrate it into EG-NAS, an evolutionary architecture search method. Evolutionary algorithms maintain and evolve a population of candidate solutions based on fitness evaluation.

Each candidate architecture is encoded by a parameter vector $\boldsymbol{\alpha} \in \mathbb{R}^{|E|\times K}$, where $E$ is the set of edges and $K$ the number of op-rule combinations. Fitness is defined by the validation accuracy after training shared weights for a few epochs using standard stochastic gradient descent (SGD).

We use CMA-ES \cite{hansen2016cma}, which evolves $\boldsymbol{\alpha}$ over time using:

\begin{equation}
    \boldsymbol{\alpha}_{t+1}^{(n)} = \mathbf{m}_t + \sigma_t \mathcal{N}(\mathbf{0}, \mathbf{C}_t),
    \quad
    \mathbf{m}_{t+1} = \sum_{n=1}^{\lambda} w_n \boldsymbol{\alpha}_{t+1}^{(n)},
\end{equation}
\begin{itemize}
    \item $\lambda$ is the number of samples per generation (population size),
    \item $\mathbf{m}_t$ is the current mean vector,
    \item $\mathbf{C}_t$ is the covariance matrix,
    \item $\sigma_t \in (0,1]$ is the step size (exploration factor),
    \item $w_n$ are recombination weights based on fitness scores.
\end{itemize}

All learning-rule-specific weights are optimized using SGD (learning rate $\eta_w \in (0,1]$), while $\boldsymbol{\alpha}$ is evolved independently using CMA-ES (step size $\eta_\alpha \in (0,1]$).

\paragraph{Search-space versatility.}
Figure~\ref{fig:search_process} visualizes the search process: (i) each layer selects a top operator (e.g., \texttt{3$\times$3 SepConv}) \emph{and} a learning rule (e.g., brSF); (ii) the resulting architecture is trained and validated; (iii) gradients or CMA-ES updates steer the next round of selection. The architecture space is expanded from $K$ to $K \times R$ choices, where $R$ is the number of available learning rules per operator.

\paragraph{Search space size.}
The original DARTS search space includes $K=8$ operations per edge. In our case, each operation is combined with $R=4$ learning rules (FA, uSF, brSF, Hebbian), resulting in $K \times R = 32$ unique op-rule pairs per edge—quadrupling the size of the original space.

\subsection{Experimental Setup}
We benchmark our work on the CIFAR10 and CIFAR100 datasets, and on the ImageNet16-120 dataset from NasBench201~\cite{dong2020nasbench201}, as well as the full ImageNet dataset. Note that for ImageNet, we do not repeat the search process and use the architecture searched for CIFAR10. We detail here the hyperparameters used for the supernetwork in Appendix 1.1. We also use code borrowed from the package Biotorch~\cite{akrout_biotorch} for the feedback alignment methods. We use Xavier initialization \cite{Glorot2010UnderstandingTD} for the weights.

\paragraph{Hardware Setup.} We use a single node that has an NVIDIA A100 GPU and a 28-core Intel Xeon CPU E5-2680 v4 and $4$ GB of RAM per core. The OS installed on the nodes is CentOS Linux version $8$.

\subsubsection{CIFAR10 and CIFAR100} 

    

Both CIFAR-10 and CIFAR-100 datasets follow the same setup. CIFAR-10 contains 50,000 training images and 10,000 test images across 10 classes, while CIFAR-100 contains 50,000 training images and 10,000 test images across 100 classes.


For both CIFAR-10 and CIFAR-100, we train the architectures resulting from the search using the following hyperparameters: batch size of 96, learning rate of 0.025, momentum of 0.9, weight decay of $3 \times 10^{-4}$, for 600 epochs. The architectures are initialized with 36 channels and have 20 layers. We enable the auxiliary tower with a weight of 0.4, apply cutout augmentation with a length of 16, use a drop path probability of 0.2 for \toolName{}-DARTS and 0.3 for \toolName{}-EG, and apply gradient clipping with a maximum norm of 5. These parameters follow the setup found in \cite{liu2018darts} and \cite{EG-NAS24}.
For the ResNet20 and ResNet56 architectures, we use the same experimental setup obtained via the hyperparameter search process done by~\cite{akrout_biotorch} on CIFAR10 and reproduce their experiments, and for retraining with the same learning rule, the same setup as the original architecture is applied.

\subsubsection{ImageNet16-120}
ImageNet16-120 \cite{chrabaszcz2017downsampled} is a subset of ImageNet \cite{deng2009imagenet} with 120 classes, downsampled to 16x16 resolution. This dataset provides a fast benchmark for evaluating the robustness of the searched architectures. The hyperparameters used to train the architectures on ImageNet16-120 are: batch size of 1024, learning rate of 0.5, momentum of 0.9, weight decay of $3 \times 10^{-5}$, for 250 epochs. The architectures are initialized with 48 channels and have 14 layers. We enable the auxiliary tower with a weight of 0.4, apply label smoothing with a factor of 0.1, use a linear learning rate scheduler, and apply gradient clipping with a maximum norm of 5. No drop path regularization is applied. These settings follow the setup described in \cite{EG-NAS24}.

\vspace{-1em}

\subsubsection{Full ImageNet}

The full ImageNet dataset \cite{deng2009imagenet} consists of 1,000 classes with high-resolution images.

The hyperparameters for training the architectures on the full ImageNet dataset are the same as the ones described in \cite{EG-NAS24}.

\subsection{Adversarial attacks}
In our experiments, we used the following parameters for the adversarial attacks, $\alpha$ denotes the step size, and $\epsilon$ the maximum perturbation magnitude.

\begin{itemize}
    \item \textbf{Fast Gradient Sign Method (FGSM):} We set the maximum perturbation size $\epsilon = 0.35$.
    \item \textbf{Projected Gradient Descent (PGD):} The step size $\alpha$ was set to $2/255$, with a maximum perturbation of $\epsilon = 0.35$ and 10 steps. We also used random initialization.
    \item \textbf{Targeted Projected Gradient Descent (TPGD):} This attack used a step size of $\alpha = 2/255$, a perturbation limit of $\epsilon = 8/255$, and 7 iterations.
    \item \textbf{One-Pixel Attack:} We altered 1 to 6 pixels, using a population size of 10 and 10 optimization steps for the differential evolution.
    \item \textbf{Auto Projected Gradient Descent (APGD):} We used the $L_\infty$ norm with a perturbation size of $\epsilon = 8/255$, 50 steps, and 1 restart.
    \item \textbf{Square AttacK \cite{andriushchenko2020square}:} A black-box $L_\infty$ attack with $\epsilon = 8/255$, which performs random search to iteratively replace square patches in the image, requiring no gradient access to the model.
    \item \textbf{Transfer Attack \cite{szegedy2013intriguing}:} We simulate a black-box setting by generating adversarial examples using FGSM ($\epsilon=8/255$) on a RobustBench surrogate model \cite{croce2021robustbench} and evaluating them on our target model.

\end{itemize}

Each attack was applied in both the untargeted and targeted modes where applicable, to explore the full range of adversarial vulnerabilities in our bio-inspired neural networks.

\section{Results}
\begin{table}[t]
\small
\centering
\footnotesize         
\begin{tabular}{p{3cm} cc}
\toprule
\textbf{Architecture} &
\multicolumn{1}{p{2cm}}{\centering\textbf{CIFAR-10\\Test Err.\;(\%)}} &
\multicolumn{1}{p{2cm}}{\centering\textbf{CIFAR-100\\Test Err.\;(\%)}} \\
\midrule
ResNet \cite{resnet16}                        & 4.61  & 22.10 \\
ENAS + cutout \cite{pham2018efficient}        & 2.89  &  –    \\
AmoebaNet-A \cite{real2019regularized-amoeba} & 3.34  & 17.63 \\
NSGA-Net \cite{lu2019nsga}                    & 2.75  & 20.74 \\
NSGANetV1-A2 \cite{lu2019nsga}                & 2.65  &  –    \\
EPCNAS-C \cite{EPCNAS}                        & 3.24  & 18.36 \\
EAEPSO \cite{eaepso_yuan_2023}                & 2.74  & 16.94 \\
DARTS (1st) \cite{liu2018darts}               & 3.00  & 17.54 \\
DARTS (2nd) \cite{liu2018darts}               & 2.76  &  –    \\
SNAS + cutout \cite{xiesnas}                  & 2.85  & 17.55 \\
ProxylessNAS + cutout \cite{cai2018proxylessnas} & \textbf{2.02} &  – \\
GDAS \cite{dong2019searching}                 & 2.93  & 18.38 \\
BayesNAS \cite{zhou2019bayesnas}              & 2.81  &  –    \\
P-DARTS + cutout \cite{pdarts}                & 2.50  & 17.49 \\
PC-DARTS + cutout \cite{chen2019progressive}  & 2.57  & 16.90 \\
DARTS- \cite{Xu2020PC-DARTS:}                 & 2.59  & 17.51 \\
$\beta$-DARTS \cite{ye2022bdarts}             & 2.53  & 16.24 \\
DrNAS \cite{chendrnas}                        & 2.54  &  –    \\
DARTS+PT \cite{wang_idarts_2021a}             & 2.61  &  –    \\
EG-NAS \cite{EG-NAS24}                        & 2.53  & \textbf{16.22} \\
\midrule
\rowcolor{gray!15} ResNet20 FA*               & 32.16 & 75.70 \\
\rowcolor{gray!15} ResNet20 frSF*             & 11.20 & 58.68 \\
\rowcolor{gray!15} ResNet20 brSF*             & 11.02 & 58.37 \\
\rowcolor{gray!15} ResNet20 uSF*              & 10.05 & 56.09 \\
\rowcolor{gray!15} ResNet56 FA*                & 34.88 &  –    \\
\rowcolor{gray!15} ResNet56 frSF*             &  9.49 & 65.66 \\
\rowcolor{gray!15} ResNet56 brSF*             &  8.69 & 58.37 \\
\rowcolor{gray!15} ResNet56 uSF*              &  8.20 & 65.23 \\
ResNet20 BP                                   &  8.63 &  –    \\
ResNet56 BP                                   &  8.30 &  –    \\
\rowcolor{gray!15} SNASNet \cite{kim-2022-nas-snn} & 5.88 & 26.96 \\
\rowcolor{gray!15} SoftHebb* \cite{moraitis-2022}& 19.60 & 43.30 \\
\rowcolor{gray!15} FastHebb \cite{lagani2022fasthebb} & 15.00 &  –    \\
\rowcolor{gray!15} \textbf{\toolName{}-EG (ours)}          &  7.16 & 29.76 \\
\rowcolor{gray!15} \textbf{\toolName{}-DARTS (ours)}      & \textbf{4.84} & \textbf{23.52} \\
\rowcolor{gray!15} \toolName{}-DARTS-HL-PC (ours) & 18.08 &  – \\
\bottomrule
\end{tabular}
\caption{\small Test error (\%) on CIFAR-10 / CIFAR-100.  
Rows shaded in gray are \emph{biologically plaus\-ible} methods; unshaded rows use standard back-prop.  
Bold numbers mark the best error within each group (bio-inspired vs.\ BP) for each dataset.  
\textbf{*} CIFAR-100 values with * were reproduced under our 150-epoch protocol.}
\label{tab:accuracies}
\end{table}

\begin{table}[h]
\centering

\begin{tabular}{l|c}
\toprule
\rowcolor{gray!20} BioNAS-DARTS (ours) & \textbf{60.51} \\
\rowcolor{gray!20} BioNAS-EG (ours) & \textbf{57.01} \\

\rowcolor{gray!20} SoftHebb     & 27.00 \\
\rowcolor{gray!20} FastHebb*  & 21.34 \\
\midrule
Inception-v1    & 69.9 \\
MobileNet       & 70.6 \\
DARTS    & 73.3 \\
SNAS            & 72.7 \\
EG-NAS          & 75.1 \\
\bottomrule
\end{tabular}
\caption{Top-1 Accuracy (\%) on ImageNet. Gray rows indicate biologically inspired models. FastHebb’s result was obtained via our code reproduction. * Result obtained from our reproduction of the experiment from the open-sourced code, since the paper reported a top-5 accuracy of 76.10\% only.}
\vspace{-2em}

\label{tab:imagenet_accuracy}
\end{table}

\begin{table*}[t]
    \centering
    \small
    \begin{tabular} {p{3cm}|p{2cm}|p{2cm}|p{2.3cm}|p{2cm}}
        \toprule
        Model & CIFAR-10 & CIFAR-100 & ImageNet 16-120 & Search Time \\
              &   Val \% &   Val \%  &      Val \%    & (GPU-Days)  \\
        \midrule
        \rowcolor{gray!20} BioNAS-EG (ours)    & 81.16 & 70.24 & 23.84 & 0.35 \\
        \rowcolor{gray!20} BioNAS-DARTS (ours) & 85.83 & 52.12 & 25.60 & 1.37 \\
        \hline
        ResNet~\cite{resnet16}  & 90.83 & 70.42 & 44.53 & - \\
        Random                  & 93.70 & 70.65 & 26.28 & - \\
        ENAS \cite{pham2018efficient}                    & 53.89 & 13.96 & 14.57 & 0.5 \\
        Random-NAS              & 84.07 & 52.31 & 26.28 & - \\
        SETN \cite{dong-yang2019one}                    & 87.64 & 59.05 & 32.52 & - \\
        DSNAS \cite{Hu_dsnas_2020}                   & 93.08 & 31.01 & 41.07 & 1.5 \\
        DARTS-V1~\cite{liu2018darts} (BP) & 54.30 & 15.61 & 16.32 & 0.4 \\

        PC-DARTS~\cite{Xu2020PC-DARTS:}               & 93.41 & 67.48 & 41.31 & 0.1 \\
        iDARTS~\cite{wang_idarts_2021a}                 & 93.58 & 70.83 & 40.8 & - \\
        DARTS~\cite{darts2020chu} & 93.80  & 71.53  & 45.12 & 0.4 \\
        EGNAS \cite{EG-NAS24}                   & 93.56 & 70.78 & 46.13 & 0.1 \\
        \bottomrule
    \end{tabular}

    \caption{Validation accuracies and search times for the search process on CIFAR-10, CIFAR-100, and ImageNet16-120 datasets. \toolName{}'s supernetwork achieves a validation accuracy of 23.84\% in the search phase and an accuracy of 43.42\% when the architecture is trained separately.}
    \vspace{-1em}

    \label{tab:accuracy_supernetwork}
\end{table*}

\begin{table*}[htbp]
\small
\centering
\begin{tabular}{l|rrrrrr}
\toprule
Attack & ResNet56-FA & ResNet56-uSF & ResNet56-frSF & ResNet56-brSF & ResNet56-BP & BioNAS-DARTS(ours) \\ \midrule
Clean   & 65.8 & 91.6 & 89.8 & 89.8 & 92.0 & \textbf{\CIFARSmall{}} \\ \midrule
OP1     & 47.8 & 63.8 & 62.6 & 65.4 & 59.6 & \textbf{90.9} \\
OP2     & 40.6 & 44.0 & 42.4 & 45.0 & 35.8 & \textbf{89.4} \\
OP3     & 38.2 & 38.4 & 40.6 & 42.0 & 32.4 & \textbf{87.5} \\
OP4     & 38.2 & 36.4 & 38.6 & 40.4 & 32.0 & \textbf{86.2} \\
OP5     & 38.0 & 34.4 & 35.4 & 38.0 & 28.2 & \textbf{84.9} \\ \midrule
FGSM    & \textbf{61.7} & 14.6 & 18.8 & 73.4 & 12.5 & 61.1 \\
PGD     & \textbf{63.3} &  0.0 &  0.0 &  0.0 &  0.0 & 60.6 \\
TPGD    & 67.2 & 40.6 & 46.1 & 50.0 & 33.6 & \textbf{67.5} \\
APGD    & 33.6 &  0.0 &  0.0 &  0.0 &  0.0 & \textbf{67.0} \\
Square  & 34.60 & 53.40 & 55.30 & 46.60 & 51.60 & \textbf{58.54} \\
Transfer Attack & 25.00 & 51.04 & 46.25 & 44.38 & 44.70 & \textbf{53.54} \\

\bottomrule
\end{tabular}
\caption{Robust accuracy (\%) of \toolName{} versus ResNet baselines under One Pixel and gradient-based attacks."OP$k$" denotes a $k$-pixel One Pixel attack. Higher values indicate better robustness. Despite using stronger attack settings in our evaluation (e.g., FGSM with $\epsilon$=0.35 vs. $\epsilon$=8/255~0.03 in RobustNAS), our framework achieves higher adversarial robustness—61.1\% under FGSM and 60.6\% under PGD—compared to RobustNAS’s optimal results of 53.2\% and 48.2\%, respectively.}
\label{tab:adversarial-robustness}
\end{table*}

We present a comprehensive evaluation of \toolName{} across multiple datasets to assess its accuracy, robustness, and the benefits of using mixed biologically-inspired learning rules.

In Table~\ref{tab:accuracies}, we compare \toolName{} to both backpropagation-based NAS methods and biologically-inspired baselines. Our model \toolName{}-DARTS achieves a test error of 4.84\% on CIFAR-10 and 23.52\% on CIFAR-100, outperforming other biologically-inspired methods such as SoftHebb and FastHebb. It also surpasses all the variants of ResNet that were trained with a single biologically inspired learning rule such as ResNet56 uSF (8.20\% on CIFAR-10), ResNet56 brSF (8.69\%), and ResNet56 frSF (9.49\%), as well as the ResNet20 and ResNet56 baselines trained with standard backpropagation (8.63\% and 8.30\%, respectively). This comparison highlights the importance of jointly searching for both architecture and learning rules, rather than fixing one in advance. We also experiment with less effective but more biologically plausible learning rules as part of the search space such as Hebbian learning and predictive coding and get an error rate reported in the table as \toolName{}-DARTS-HL-PC of 18.08 on CIFAR-10, which is an acceptable accuracy knowing that the search tended to pick some Hebbian layers as part of the final architecture.

We note that both SoftHebb and FastHebb are self-supervised approaches, which makes a direct comparison with our supervised setting less straightforward. However, they offer valuable insight into the diversity of bio-inspired models. Our method shows that mixing rules in a supervised NAS framework offers higher accuracy compared to using any single biologically-inspired rule. It also shows that mixing learning rules improves the performance of bio-inspired neural networks to the point where they reach an accuracy comparable to neural networks trained with BP.

Table~\ref{tab:imagenet_accuracy} extends this analysis to ImageNet. While DARTS and other conventional models trained with BP still lead in accuracy, our BioNAS-DARTS achieves a 60.51\% top-1 accuracy, significantly outperforming SoftHebb (27.00\%) and FastHebb (21.34\%, as reproduced). Though lower than traditional methods like DARTS (73.3\%) or EG-NAS (75.1\%).

As seen in Table~\ref{tab:accuracy_supernetwork}, our approach achieves competitive results within a modest compute budget (e.g., 0.35 GPU-days for \toolName{}-EG). On ImageNet-16-120, when we train the architecture searched on CIFAR-10 we obtain an accuracy of \ImageNetSmall{} despite having different spatial resolutions. This demonstrates the practical efficiency of our approach compared to other NAS frameworks.

\subsection{Adversarial Robustness}

Table~\ref{tab:adversarial-robustness} compares the adversarial robustness across various models and training methods. The rows correspond to different attack types: \textit{Clean} denotes accuracy on unperturbed data; OP1 through OP5 represent One Pixel attacks modifying 1 to 5 pixels respectively (a type of black-box attack); and the bottom rows include gradient-based white-box attacks: FGSM, PGD, TPGD, and APGD. The columns compare several ResNet-56 variants trained with individual bio-inspired learning rules (FA, uSF, frSF, brSF), a ResNet-20 trained with standard backpropagation (RN56-BP), and our \toolName{}-DARTS model trained with mixed learning rules.

On clean data, \toolName{}-DARTS achieves the highest accuracy (\CIFARSmall{}\%), exceeding all baselines. Under black-box One Pixel attacks (OP1--OP5), it consistently outperforms the others, retaining high accuracy between 90.9\% (OP1) and 84.9\% (OP5), whereas all other models degrade significantly in performance as the number of perturbed pixels increases. In contrast, standard backpropagation and single-rule models like RN56-uSF and RN20-BP drop to accuracies as low as 28--44\%.
For the square attack (black box with no access to gradients), our framework outperforms all existing baselines with a robust precision of 58. 54\% compared to 55. 30\% for the best RN, as shown in Table \ref{tab:adversarial-robustness}.

For gradient-based attacks, results are more nuanced. \toolName{}-DARTS achieves top performance under APGD (67.0\%) and TPGD (67.5\%), while brSF and FA perform best under FGSM and PGD, respectively. Notably, many models trained with uSF, frSF, and BP collapse to 0\% accuracy under stronger attacks like PGD and APGD, while \toolName{}-DARTS maintains robust performance.

These results illustrate that mixing learning rules during NAS search yields models with significantly stronger resilience to adversarial attacks, especially in the more challenging black-box setting.

Finally, our framework offers a promising path forward for those desiring more biological plausibility in deep learning models. It provides an automated method for combining architectural search with bio-inspired rules and is particularly useful for researchers modeling brain functions.
\vspace{-1em}

\subsection{Discussion}

Our experiments demonstrate that incorporating mixed learning rules into the search space improves model accuracy and robustness.

Our NAS framework assigns different bio-inspired learning rules to different layers. A natural question that arises is whether the best combination of learning rules found by our framework always follows a particular pattern (for example, using particular rules in particular layers always). To answer this question, we randomly assigned different bio-inspired learning rules to different layers of the neural network. Table~\ref{tab:compare-lr-darts} shows that the accuracy remains high after such random assignment of learning rules. This shows that the benefit of mixing rules is not due to the use of a particular pattern or combination of learning rules.

While some combinations of learning rules perform slightly better than others, the differences are minimal, indicating that mixed learning rules provide a stable and effective training mechanism. We also provide a preliminary analysis in Appendix 1.2 where our observations suggest that mixing learning rules yields behaviors analogous to L1 and L2 regularization, potentially promoting smoother or more stable learning. However, a complete theoretical formulation and proof is left for future work.



\begin{table}[t]
    \centering
    \small
    \begin{tabular}{c|c}
        \toprule
        Model & Top-1 accuracy \% \\ \hline
        BioNAS-DARTS & 94.86 \\
        BioNAS-DARTS1 & 94.87 \\
        BioNAS-DARTS2 & 94.88 \\
        BioNAS-DARTS3 & 94.83 \\
        BioNAS-DARTS6 & 92.78 \\ 
        \bottomrule
    \end{tabular}
    \vspace{-1em}

    \caption{Accuracy across random learning rule assignments in BioNAS-DARTS.}
    \vspace{-1em}

    \label{tab:compare-lr-darts}
\end{table}





\subsection{Gradient Variance Analysis}

To further understand the impact of integrating multiple biologically inspired learning rules into NAS, we conducted an empirical gradient variance analysis. We evaluated how the variance of the gradients evolves when training the architecture discovered by BioNAS-DARTS under different configurations: using a single learning rule across all layers versus using a mixed configuration as discovered during search.

\begin{figure}
    \centering
    \includegraphics[width=0.9\linewidth]{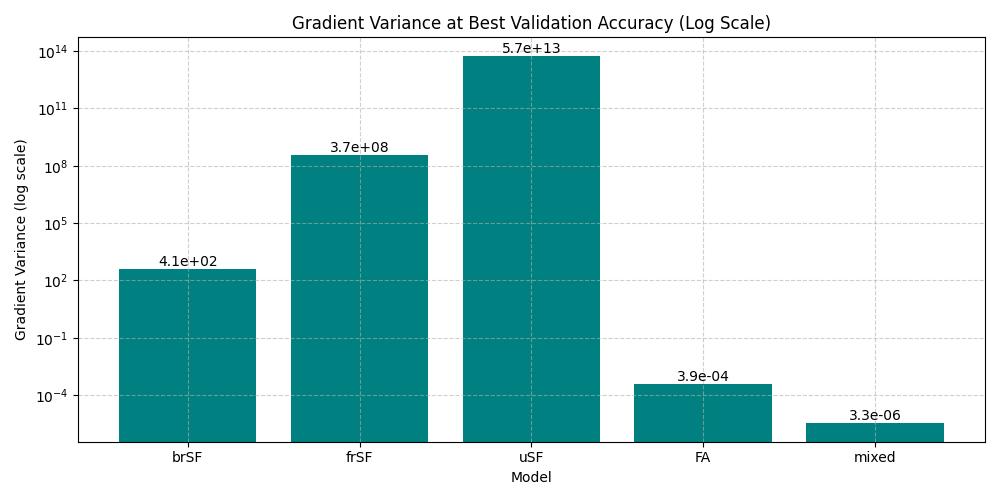}
    \vspace{-1em}

    \caption{Gradient variance over epochs when training BioNAS-DARTS with different learning rule configurations (log scale).}
    \label{fig:gradient-variance}
\end{figure}

Our results, illustrated in Figure~\ref{fig:gradient-variance}, show that BioNAS-DARTS trained with mixed learning rules exhibits consistently lower gradient variance compared to training with a fixed rule. Gradient variance has been discussed in the literature as a factor that can influence convergence speed and stability in optimization \cite{bottou2018optimization, robbins1951stochastic}. 
The lower gradient fluctuations observed in our NAS framework might help preserve effective learning dynamics, especially when using biologically plausible feedback mechanisms. While our findings are preliminary and further theoretical investigation is needed, this analysis offers an additional perspective into the effects of mixed learning rules, motivating further research in this area.

\vspace{-2em}


\section{Conclusion}
In this paper, we show that a good choice of the learning rule used for each layer in a network and the operation can result in a bio-inspired neural network with a good accuracy competing with backpropagation-trained neural networks, achieving state-of-the-art bio-inspired neural network accuracy. The mixed-rule approach provides advantages over using a single learning rule throughout the network by reducing gradient variance, facilitating better exploration of the loss landscape, and enhancing adversarial robustness. This theoretical framework aligns with the empirical results observed in our experiments, where architectures searched using NAS with mixed learning rules outperform those using a uniform rule across all layers. Further investigation should be done on the learning dynamics in this searched architecture and the representations learned. 
\vspace{-1em}

\section{Acknowledgment}This research has been partly supported by the Center for Artificial Intelligence and Robotics (CAIR) at New York University Abu Dhabi, funded by Tamkeen under the NYUAD Research Institute Award CG010 and by the Center for Cyber Security (CCS) at New York University Abu Dhabi. The research was carried out on the High-Performance Computing resources at New York University Abu Dhabi.





\bibliography{ecai-template/main}
\section{Appendix}
\subsection{Supernetwork Hyperparameters}
Table \ref{tab:training_hyperparameters}  summarizes the hyperparameters used for the supernetworks training for CIFAR10 and CIFAR100. When it comes to ImageNet16-120, an initial learning rate of 0.5 is used, and a batch size of 512.
\begin{table}[ht]
    \centering
    \begin{tabular}{ll}
        \toprule
        \textbf{Hyper-parameter} & \textbf{Value} \\
        \midrule
        Optimizer        & SGD \\
        Initial LR       & 0.1 \\
        Nesterov         & Yes \\
        Ending LR        & 0.0 \\
        Momentum         & 0.9 \\
        LR Schedule      & (Not mentioned) \\
        Weight Decay     & 0.0003 \\
        Epochs           & 50 \\
        Batch Size       & 256 \\
        Initial Channels & 16 \\
        V (Nodes per Cell) & 4 \\
        N (Cells)        & 5 \\
        Random Flip      & p=0.5 \\
        Random Crop      & Yes \\
        Normalization    & Yes \\
        \bottomrule
    \end{tabular}
    \caption{Hyperparameters used for training the supernetwork on CIFAR10 and CIFAR100.}
    \label{tab:training_hyperparameters}
\end{table}

\subsection{Weight distribution analysis} \label{weightdistribution}
\begin{figure}
    \centering
    \includegraphics[width=0.95\linewidth]{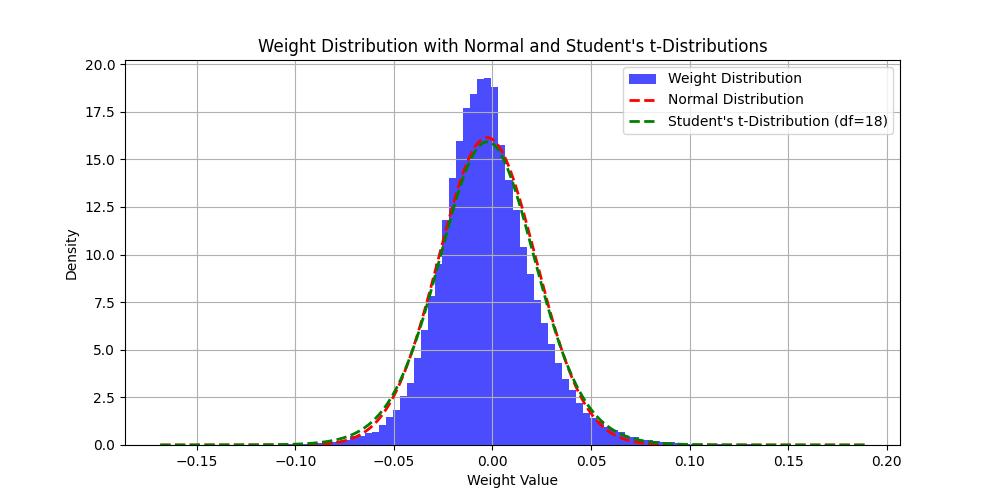}
    \caption{Weight distribution of \toolName{}-EG compared to a gaussian and and a student distribution with 10 degrees of freedom.}
    \label{fig:weight_distribution_egnas}
\end{figure}

To understand how the learning happens, we plot the weight distribution of our model as shown in Figure \ref{fig:weight_distribution_egnas}, the distribution is a little bit steeper than a Gaussian which is an effect observed usually with regularization.

\begin{figure}
     \centering
     \includegraphics[width=0.95\linewidth]{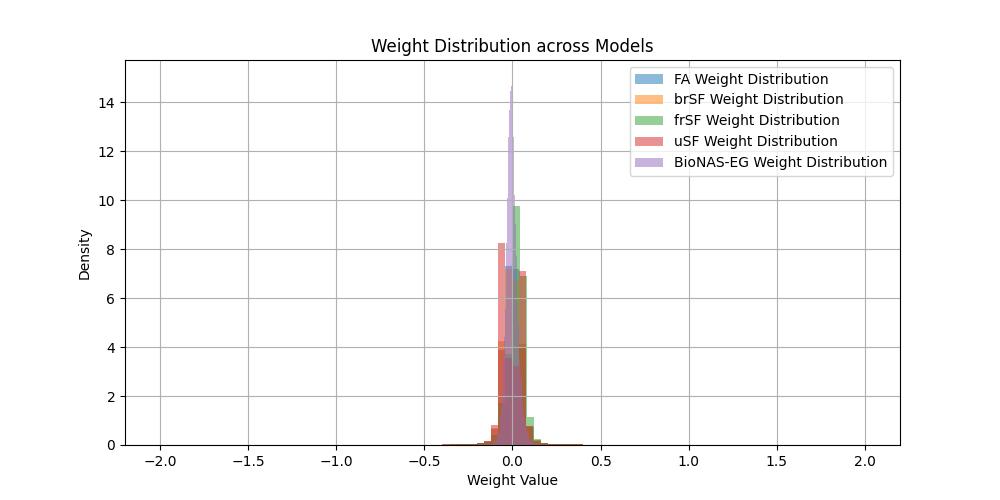}
     \caption{Weight distribution comparison between \toolName{}-EG vs training it with the same learning rule (FA, frSF, brSF, uSF) from end to end. }
     \label{fig:best_model_weights}
 \end{figure}

We compare the weights of the models with the highest accuracies, we train the resulting \toolName{}-EG architecture with the same learning rule from end to end, as shown in figure \ref{fig:best_model_weights}, the weights of \toolName{}-EG is steeper with more weights around 0. Note that we drop the weights that are less than 2 and more than 2 since their number is negligible and invisible on the plot.

\begin{figure}[ht]
    \centering
    \begin{minipage}{0.48\textwidth}
        \centering
        \includegraphics[width=\linewidth]{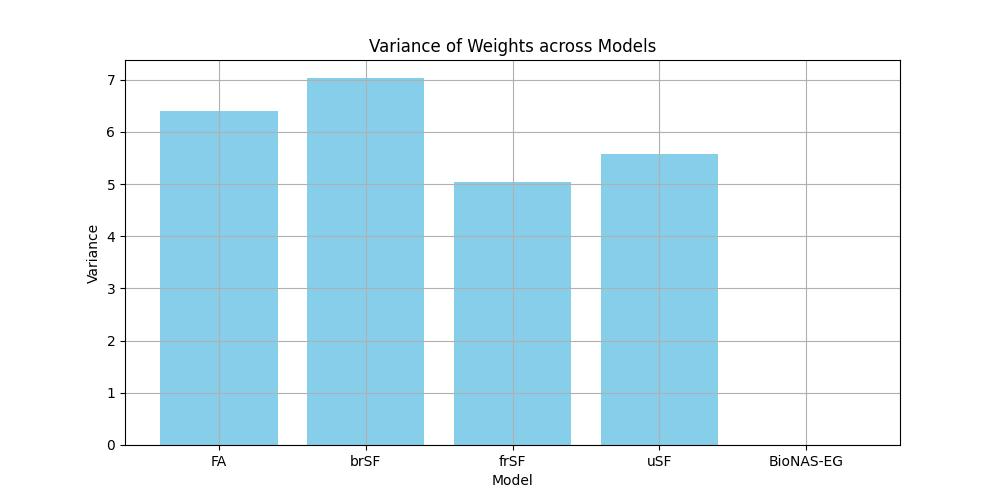}
        \label{fig:bioegnas-variance}
    \end{minipage}\hfill
    \begin{minipage}{0.49\textwidth}
        \centering
        \includegraphics[width=\linewidth]{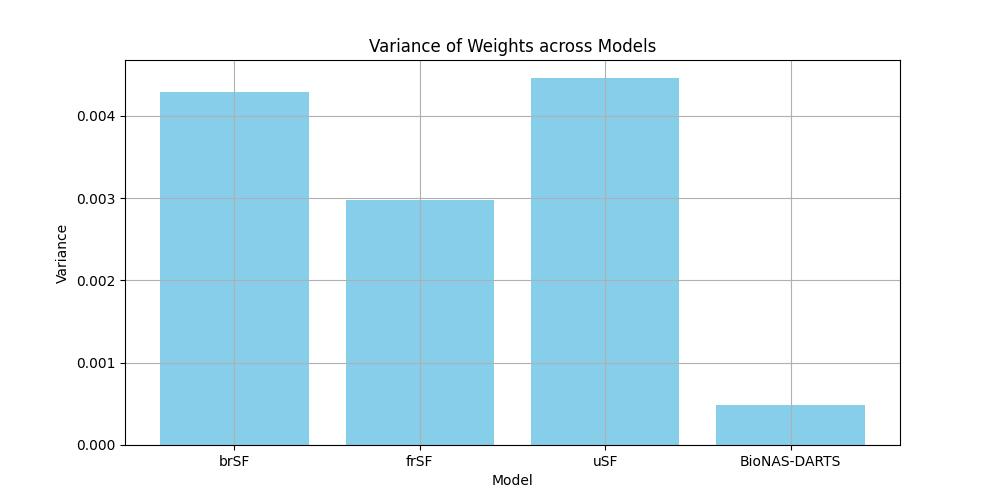}
        \label{fig:biodarts-variance}
    \end{minipage}
    \caption{Comparison of weight variance and performance between \toolName{}-EG (left) and \toolName{}-DARTS (right) models (with mixed learning rules) versus training the same resulting architecture with the same learning rule from end-to-end.}
    \label{fig:variance}
\end{figure}

 Another thing we noticed is that for training with the same learning rule, the variance is much smaller than with \toolName{}-EG as illustrated in figure \ref{fig:variance}.

\subsection{Code Implementations}

\subsubsection{Hebbian Convolution}
The Hebbian convolution's implementation is as follows:

\begin{lstlisting}[language=Python, caption=Hebbian Convolution Implementation]
class HebbianConv(nn.Module):
    def __init__(...):
        ...
        self.hebbian_weights = torch.       
        zeros_like(self.conv.weight,
        
        requires_grad=False)
        self.hebbian_scale = 1e-4

    def forward(self, x):
        # Compute Hebbian weight update
        ...
        self.hebbian_weights.   
        
        add_(hebbian_update)
        self.conv.weight.data.copy_
            
        (self.hebbian_weights)
        ...
\end{lstlisting}

\subsubsection{Predictive Coding Convolution}
The Predictive Convolution Operation's implementation is as follows:
\begin{lstlisting}[language=Python, caption=Predictive Coding Convolution Implementation]
class PredictiveCodingConv(nn.Module):
    def __init__(...):
        ...
        self.error_weights = nn.
        Parameter(torch.ones(
        out_channels),
        requires_grad=True)
        if self.gating:
            self.gate = nn.Sequential(
                nn.Conv2d(out_channels,  
                out_channels, 
                kernel_size=1, stride=1, 
                bias=False),
                nn.Sigmoid()
            )

    def forward(self, x):
        pred = self.conv(x)
        for _ in range(self.
        
        prediction_steps):
            error = x - pred
            error_update = self.
            error_weights.view(1, -1,
            1, 1) * error
            pred += error_update
        ...
\end{lstlisting}

\subsection{Theoretical Justification of Mixed Learning Rules}

To demonstrate why a neural architecture searched with NAS using mixed learning rules outperforms architectures that use the same learning rule across all layers, we analyze the impact of combining various learning rules on gradient propagation, variance reduction, and optimization efficiency. 

\subsubsection{Gradient Propagation and Learning Dynamics}

Consider a neural network consisting of multiple layers, each trained with different feedback learning rules. The general weight update in a layer \(i\) can be written as:

\begin{equation}
    \delta W_i = -(B_{i+1} \delta \mathbf{z}_{i+1}) \phi'(\mathbf{z}_i) \mathbf{y}_{i-1},
\end{equation}
where \(B_{i+1}\) is the feedback matrix and \(\delta \mathbf{z}_{i+1}\) is the gradient of the loss with respect to the input of layer \(i+1\).

When a network uses a single learning rule throughout, this feedback matrix \(B_{i+1}\) is updated uniformly across all layers. However, when different learning rules are applied in different layers, each rule adapts the feedback matrix differently, allowing for a broader range of gradient dynamics across the network. This diversity in updates helps mitigate issues like gradient vanishing or explosion, promoting better gradient flow.

\subsubsection{Exploration and Optimization Efficiency}

Each learning rule contributes differently to the exploration of the loss landscape. The variation across layers helps the network strike a balance between exploration and exploitation during optimization. For example, some rules might encourage the network to explore different parts of the loss surface by generating different gradient directions, while other rules help converge efficiently by stabilizing the updates. 

The overall gradient update at time \(t\) for a mixed-rule network is:

\begin{equation}
    \Delta \theta_t^\text{Mixed} = \sum_{k} w_k \Delta \theta_t^k,
\end{equation}
where \(\Delta \theta_t^k\) is the gradient update under rule \(k\) and \(w_k\) is the weight assigned to that rule. By combining these updates, the mixed-rule network benefits from diverse gradient dynamics that prevent overfitting and improve convergence.

\subsubsection{Adversarial Robustness}

Using mixed learning rules enhances adversarial robustness by disrupting the consistency in gradient signals across layers. In adversarial attacks, the goal is to align perturbations with the network’s gradient, but with different learning rules in different layers, the adversary faces a moving target. This makes it harder to craft perturbations that generalize across the network, improving resistance to adversarial examples.
\subsection{Neural Architecture Search}
\subsubsection{DARTS}
\label{appendix_darts}
In DARTS, a continuous relaxation of the architecture search space is employed, enabling the use of gradient-based optimization to search for the optimal architecture. Specifically, each intermediate node is computed by applying a softmax function over a mixture of candidate operations:

\begin{equation}
    o^{(i,j)}(x^{(i)}) = \sum_{o \in O} \frac{\exp(\alpha_o^{(i,j)})}{\sum_{o' \in O} \exp(\alpha_{o'}^{(i,j)})} o(x^{(i)}),
\end{equation}

where \( i < j \), the set of candidate operations is represented by \( O \), and \( \alpha_o^{(i,j)} \) represents the mixing weight for operation \( o^{(i,j)} \) in the supernetwork. During the search process, DARTS optimizes both the network weights \( \omega \) and the architecture parameters \( \alpha \) simultaneously using a bi-level optimization framework. This framework is defined as:

\begin{equation}
    \min_{\alpha} F(\omega^*(\alpha), \alpha) = L_{\text{val}}(\omega^*(\alpha), \alpha)
\end{equation}
\begin{equation}
    \text{subject to } \omega^*(\alpha) = \arg\min_{\omega} L_{\text{train}}(\omega, \alpha),
\end{equation}

where \( \omega^*(\alpha) \) denotes the optimal network weights for a given architecture. Both \( \alpha \) and \( \omega \) are updated through gradient descent. After the search phase concludes, the final architecture is derived by selecting the operation with the highest architectural parameter \( \alpha \) on each edge:

\begin{equation}
    o^{(i,j)} = \arg\max_{o \in O} \alpha_o^{(i,j)}.
\end{equation}

\subsubsection{EG-NAS}
\label{appendix_egnas}

In traditional NAS approaches, gradient-based methods typically calculate gradients and use them to guide the search for optimal architectures. However, these methods are often limited by the gradient's inherent directionality, leading to a potential risk of getting trapped in local minima. EG-NAS addresses this issue by combining Covariance Matrix Adaptive Evolutionary Strategy (CMA-ES) with gradient-based optimization. CMA-ES is a powerful evolutionary algorithm known for its global search capabilities and efficient convergence in black-box optimization problems~\cite{hansen2016cma,loshchilov2016cma}.

The EG-NAS algorithm begins by sampling $N$ architectures, $\{x_n\}_{n=1}^N$, from a Gaussian distribution with a mean vector $\alpha$ and covariance matrix $I$, such that:

\begin{equation}
\begin{aligned}
    x_n &= \alpha + \sigma y, \quad y \sim \mathcal{N}(0, I), \\
    n &= 1, 2, \dots, N.
\end{aligned}
\end{equation}
Each sampled architecture $x_n$ initializes a CMA-ES search, where $x_n$ serves as the mean vector $m_0$ of the $n$-th search. The initial population of the $n$-th evolutionary search is sampled as:

\begin{equation}
    z_i^t = m^t + \sigma y_i, \quad y_i \sim \mathcal{N}(0, C_t),
\end{equation}

where $t$ represents the iteration index, starting from $t=0$, with $C_0 = I$ and $\sigma$ as the step size. During each iteration, the mean vector $m^{t+1}$ and covariance matrix $C_{t+1}$ are updated using the following equations:

\begin{equation}
    m^{t+1} = \sum_{i=1}^{\lfloor \lambda/2 \rfloor} \beta_i z_i^t,
\end{equation}

\begin{equation}
\begin{aligned}
C_{t+1} &= (1 - c_1 - c_{\lfloor \lambda/2 \rfloor}) C_t 
+ c_1 (pp^T) \\
&\quad + c_{\lfloor \lambda/2 \rfloor} \sum_{i=1}^{\lfloor \lambda/2 \rfloor} 
\beta_i (z_i^t - m^{t+1})(z_i^t - m^{t+1})^T,
\end{aligned}
\end{equation}
where $\beta_i$ represents the fitness weight for each individual, and $p$ is the evolutionary path. 

Each individual is evaluated by a compound fitness function, incorporating both the cross-entropy loss ($L_1$) and cosine similarity ($L_2$):

\begin{equation}
    f(\alpha^t, z_i^{t+1}) =
\begin{cases} 
\zeta L_1 - \eta L_2 & \text{if } \text{Acc}(\alpha^t) > \text{Acc}(z_i^{t+1}) \\
\zeta L_1 + \eta L_2 & \text{else},
\end{cases}
\end{equation}

where $\zeta$ and $\eta$ are weight coefficients, and $L_2$ ensures diversity among architectures by calculating the cosine similarity between $\alpha^t$ and $z_i^{t+1}$. This compound fitness function helps avoid premature convergence and ensures exploration of diverse architectures.

Finally, EG-NAS integrates the evolutionary strategy with gradient-based optimization. The network weights $\omega$ are updated via gradient descent, while the architecture parameters $\alpha$ are updated using the evolutionary strategy. The update rule for $\alpha$ is:

\begin{equation}
    \alpha^t = \alpha^{t-1} + \xi s^t,
\end{equation}
where $s^t$ represents the search direction based on fitness values, and $\xi$ is the step size. The best-performing architecture across all evolutionary steps is selected as the final output, ensuring a more diverse and effective architecture search than traditional gradient-based methods.

\end{document}